\documentclass[conference]{IEEEtran}
\IEEEoverridecommandlockouts
\usepackage{cite}
\usepackage{amsmath,amssymb,amsfonts}
\usepackage{algorithmic}
\usepackage{blindtext,graphicx}
\usepackage[absolute]{textpos}
\usepackage{textcomp}
\usepackage{xcolor}
\def\BibTeX{{\rm B\kern-.05em{\sc i\kern-.025em b}\kern-.08em
    T\kern-.1667em\lower.7ex\hbox{E}\kern-.125emX}}
\begin{document}

\begin{textblock}{15}(2,1)
\noindent\footnotesize ©2019 IEEE.  Personal use of this material is permitted.  Permission from IEEE must be obtained for all other uses, in any current or future media, including reprinting/republishing this material for advertising or promotional purposes, creating new collective works, for resale or redistribution to servers or lists, or reuse of any copyrighted component of this work in other works.
\end{textblock}

\title{Autonomous Reinforcement Learning of\\ Multiple Interrelated Tasks\\
\thanks{This project has received funding from the European
Union’s Horizon 2020 Research and Innovation Program under
Grant Agreement no. 713010 (GOAL-Robots – Goal-based
Open-ended Autonomous Learning Robots).}
}

\author{\IEEEauthorblockN{1\textsuperscript{st} Vieri Giuliano Santucci, 2\textsuperscript{nd} Gianluca Baldassarre, 3\textsuperscript{rd} Emilio Cartoni}
\IEEEauthorblockA{\textit{Istituto di Scienze e Tecnologie della Cognizione (ISTC)} \\
\textit{Consiglio Nazionale delle Ricerche (CNR)}\\
Rome, Italy \\
\{vieri.santucci,gianluca.baldassarre,emilio.cartoni\}@istc.cnr.it}
}

\maketitle

\begin{abstract}
Autonomous multiple tasks learning is a fundamental capability to develop versatile artificial agents that can act in complex environments. In real-world scenarios, tasks may be interrelated (or ``hierarchical'') so that a robot has to first learn to achieve some of them to set the preconditions for learning other ones. 
Even though different strategies have been used in robotics to tackle the acquisition of interrelated tasks, in particular within the developmental robotics framework, autonomous learning in this kind of scenarios is still an open question. 
Building on previous research in the framework of intrinsically motivated open-ended learning, in this work we describe how this question can be addressed working on the level of task selection, in particular considering the multiple interrelated tasks scenario as an MDP where the system is trying to maximise its competence over all the tasks.
\end{abstract}


\section{Introduction}
Autonomy, intended as the capability of a system to behave and learn without pre-assigned tasks or externally provided knowledge (i.e. human designers knowledge), is a paramount challenge for the development of artificial agents that can act in complex and continuously changing real-world scenarios. Acquiring many different skills is the necessary starting point to foster versatility and adaptation, thus autonomous open-ended learning of skills can be considered one of the main topics for research in robotics. While the learning of multiple skills 
can be addressed through different machine learning techniques by sequentially assigning to the robot a series of $N$ tasks, autonomy implies that the agent has the capability to select on which task to focus and shift between them possibly in a smart way.    

Intrinsic motivations (IMs) have been used in the field of machine learning and developmental robotics \cite{Oudeyer2007intrinsic,Schmidhuber2010,Baldassarre2013Book} to provide self-generated reinforcement signals driving exploration and skill learning \cite{Santucci2014,Hafez2017,Dhakan2018,Tanneberg2019}. Other studies \cite{Baranes2013,Santucci2016,Forestier2017} implemented IMs as a motivational signal for the autonomous selection of tasks. Some tasks can be defined in terms of ``goals'', that is desirable environment states the agent might aim to accomplish (here we focus on this type of tasks and so we use the terms ``goal'' and ``task'' interchangeably). 

The learning progress in achieving a goal can be used as a transient reward so that the system focuses on tasks where it is learning the most, moving to other ones when the task-related skill has been completely learnt or when more promising activities come at hand \cite{Lopes2012,Santucci2013best}. This strategy allows the learning of multiple separated skills, and possibly a dynamical transfer of knowledge between tasks that require similar policies \cite{Seepanomwan2017}.

In real-world scenarios goals may need specific initial conditions to be performed; or they may be interrelated so that to achieve a task the robot needs first to learn and accomplish other ones. This last case is of particular interest and although it has been studied under different headings it is still an open issue from an autonomous open-ended learning perspective. 

Hierarchical reinforcement learning \cite{Barto2003} has been combined with IMs to allow for the autonomous formation of skills sequences, but usually these methods tackle only discrete states and actions domains \cite{Vigorito2010}; or focus on the discovery of sub-goals on the basis of externally given tasks \cite{Bakker2004}; or under the assumption that sub-goals come as predefined rewards \cite{Niel2018}, thereby reducing the autonomy of the agent during the learning process. Trajectories with via points \cite{Reinhart2017} and parametrised skills \cite{DaSilva2014}  are able to learn multiple motor trajectories, but this is commonly done  considering single tasks/skills or assuming pre-defined tasks. Imitation learning has achieved important results in the learning of task hierarchies \cite{Grollman2010,Mohseni2018,Nair2018}, even in association with IMs \cite{Duminy2018}, but by definition it relies on external knowledge sources (i.e. the ``instructor'') thus limiting the autonomy of the systems.

Without considering other important issues for life-long open-ended learning such as the autonomous discovery \cite{Rolf2014,Santucci2016,Meeden2017} or the autonomous generation of tasks/goals \cite{Cartoni2018}, in this paper we describe how learning multiple interrelated tasks can be tackled focusing on the level of task selection, providing an analysis of the problem together with the proposed solution (Sec.~\ref{sec:ProblemDescription}). Moreover, we test our hypothesis by comparing different systems (Sec.~\ref{sec:M-GRAIL}), implemented as enhancements of the GRAIL architecture \cite{Santucci2016}, in a simulated robotic scenario involving multiple interrelated tasks (Sec.~\ref{sec:Setup}).   

\section{Description of the problem and proposed solution}
\label{sec:ProblemDescription}

From a reinforcement learning (RL) perspective \cite{Sutton1998}, the learning of multiple goals can be seen as the learning of different policies $\pi_g$, each one associated to a different goal $g \in G$. Those policies maximise the return provided by the reward function $R_g$ associated with goal $g$ (see also \cite{Florensa2018, Kulkarni2016}). For each $g$ the system thus aims to learn a policy: 
\begin{equation}\label{eqn:objFuncMDP}
   \pi_g^*(a|s) = \underset{\pi}{\text{argmax}} \, R_g(\pi_g)
\end{equation}

Since we are considering an open-ended learning scenario where no specific task is assigned to the robot, we suppose the system is not maximising extrinsic rewards, but rather a competence function $C$ over the distribution of goals $G$. Here, $C$ is the sum of the agent's competence $C_g$ at each goal $g$ as made possible by a given candidate goal-selection policy $\Pi_t$. In other words, the overall competence is a measure of the agent's ability to efficiently accomplish different goals by allocating its learning time among them using a given policy $\Pi_t$ associated with an MDP where the agent is learning to maximise the competence $C_g$ for that goal rather than the specific reward $R_g$ for that goal. If we consider a finite time horizon $T$, the robot needs to properly allocate its training time to the goals that guarantee the highest competence gain. To do so, the system may use the current derivative of the competence $\delta C$ (w.r.t. time) as an intrinsic motivation signal to select the goal with the highest competence improvement at each time step $t$, where time here refers to one training step over a given task (the efficacy of this approach has been shown in different works within the intrinsically motivated open-ended learning framework \cite{Lopes2012,Santucci2013iccm,Merrick2012}). The problem of task selection can thus be described as an $N$-armed bandit \cite{Sutton1998} (possibly a \textit{rotting bandit} \cite{Levine2017} due to the non-stationary transient nature of IMs) where the agent learns a policy $\Pi$ to select goals that maximise the current competence improvement $\delta C$:
\begin{equation}\label{eqn:objFuncBandit}
   \Pi^* =  \underset{\Pi}{\text{argmax}} \ \delta C (\Pi_t)
\end{equation}

If we constrain the feasibility of the goals to specific environmental conditions, goal selection becomes a \textit{contextual bandit} problem \cite{Sutton1998} where the robot has to learn the value of goals and selects them depending on the current state $s \in S$. Equation \ref{eqn:objFuncBandit} thus becomes:
\begin{equation}\label{eqn:objFuncContestualBandit}
   \Pi^*(s_t) =  \underset{\Pi}{\text{argmax}} \ \delta C (\Pi(s_t))
\end{equation}
\noindent where now the policy for selecting goals for training needs to explicitly take into account the current state of the agent, which may include information such as which other goals have already been accomplished. By making this change to the objective, the system can bias the choice using the expected competence gain for each goal given different conditions. The evaluation of the competence improvement for each goal can be done via a state-based moving average of performance at achieving that goal given the current policy. 

If we now further assume a situation where goals are \textit{interrelated}, so that a goal may be a precondition for other ones, we shift to a different kind of problem where the state of the environment depends on previously selected (and possibly achieved) goals. A sequence of contextual bandits where the context at time $t+1$ is determined by the action (here goal selection) executed at time $t$, can be seen as an MDP over all the goal-specific MDPs for which the robot is learning the policy (a ``skill''). This is the typical situation of hierarchical skill learning that is still scarcely addressed within a fully autonomous open-ended framework.

What we propose is that, given the structure of the problem, goal selection for multiple interrelated tasks can be treated as an MDP and, consequently, can be addressed via RL algorithms that transfer \textit{intrinsic-motivation} values between interrelated goals. In particular, in the following sections we show how a system implementing goal selection with a standard Q-learning algorithm \cite{Sutton1998} is able to outperform systems that treat it as a standard bandit or contextual bandit problem.

\begin{figure}
    \centering
    \includegraphics[width=.6 \columnwidth,keepaspectratio]{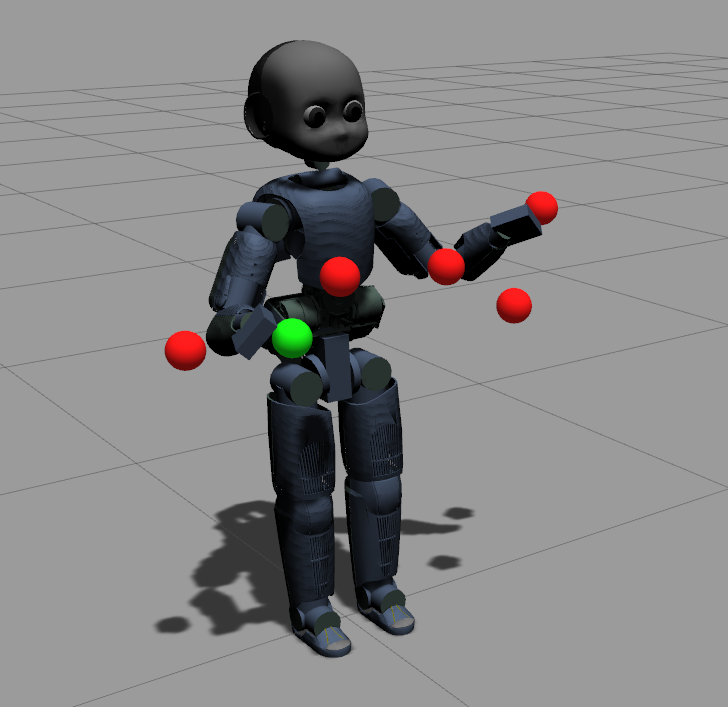}
    \caption{The simulated iCub in the experimental setup. When a sphere is touched (given its preconditions) it ``lights up'' changing its colour to green.}
    \label{fig:Robot}
\end{figure}

\section{Methods: setup and system}
\label{sec:Methods}

\subsection{The robot and the experimental scenarios}
\label{sec:Setup}

To test our hypothesis we designed an experimental scenario where a simulated iCub robot \cite{Metta2008} has to perform multiple interrelated tasks consisting in touching-to-activate different spheres anchored to the world (the spheres “float” in front of the robot). Notwithstanding its simplicity, this setup allow us to test all the issues related to goal-selection and skill learning of multiple interrelated tasks.
We use the two arms of the iCub robot each with 4 degrees-of-freedom (DOFs) while the joints of the wrists are kept fixed and hands are substituted with 2 scoops. Collisions are disabled in the simulator while a sensor in the centre of each scoop determines whether the robot touches one of the spheres. Some spheres may be ``conditioned'' to other ones, so that to activate one of them the robot has to previously activate other spheres. In this way if the system wants to learn a skill, it has to set the environment in the proper condition, i.e. it has to first select and achieve other goals that constitute the precondition for the one to be trained.

We designed different variations of the general setup to better show the advantages of our solution. In particular we run three different experiments:

\begin{enumerate}
    \item \textit{No relations / N-armed bandit:} all the spheres (here six) can be activated independently from the state of the environment and from the other spheres.
    \item \textit{Environmental dependence / Contextual bandit}: the activation of a sphere, by having the robot touch it, is dependent on some environmental variable. In this setting we assume a state feature (the ``contextual feature'') that is set to 1.0 with 50\% probability at the beginning of each trial, and to 0.0 otherwise. The six spheres the systems is learning to achieve here depend on the context: three of them can only be activated when the contextual feature is on, and the other three only when it is off.
    \item \textit{Multiple Interrelated Tasks / MDP}: the ``achievability'' of a task (activation of a sphere) is now dependent on the activation status of the other spheres. In this scenario, the fact that the robot has previously achieved or not a goal (or set of goals) constitutes the precondition for the achievement of other goals, thus introducing interdependencies between the available tasks (see Sec.~\ref{sec:ThirdExperiment} for details).
\end{enumerate}

\subsection{Compared systems}
\label{sec:M-GRAIL}

All the compared systems are developed building on a previous architecture called GRAIL \cite{Santucci2016}, developed for the autonomous discovery and intrinsically motivated learning of goals. Due to paper length constraints here we describe only the features of GRAIL (and the modifications proposed in the current work) that are useful for the understanding of the presented results, and we invite readers to refer to the cited work for further details.

GRAIL has a high-level component, the goal selector (GS), determining at each trial the task the system is pursuing (see Sec.~\ref{sec:Results} for a description of the experimental scheduling). The selected goal is then used by the expert selector (ES) to choose with which module (the expert) to learn the low-level control policy (the skill) for that task. There are two experts associated with each goal: for each goal, the ES can choose one between them to control either one of the two arms of the robot. This gives a higher versatility to the robot (see \cite{Santucci2014icdl}), but in this work we do not focus on this aspect. At the lower-level of skill learning (training of the policy), any implementation could be used. In GRAIL we developed each expert as an actor-critic network modified to work with continuous state and action spaces \cite{Doya2000} and trained through a TD-Learning algorithm on the basis of the pseudo-reward signals generated for achieving the selected goal (lighting up the target sphere). At every time step the selected expert receives as input the angles of the four actuated joints of the arm and returns as output four desired joint angles to move the arm through position control.

The autonomous selection of the goals is performed by the GS according to a competence-based intrinsic motivation signal (CB-IM) calculated over each goal (see \cite{Santucci2013best} for the comparison of different types of IMs). In particular, the CB-IM signal of a goal is the competence prediction improvement (CPI) of a predictor that receives as input the selected goal and produces as output the predicted probability of the achievement of that goal within the trial.

\begin{figure}
    \centering
    \includegraphics[width=.85 \columnwidth,keepaspectratio]{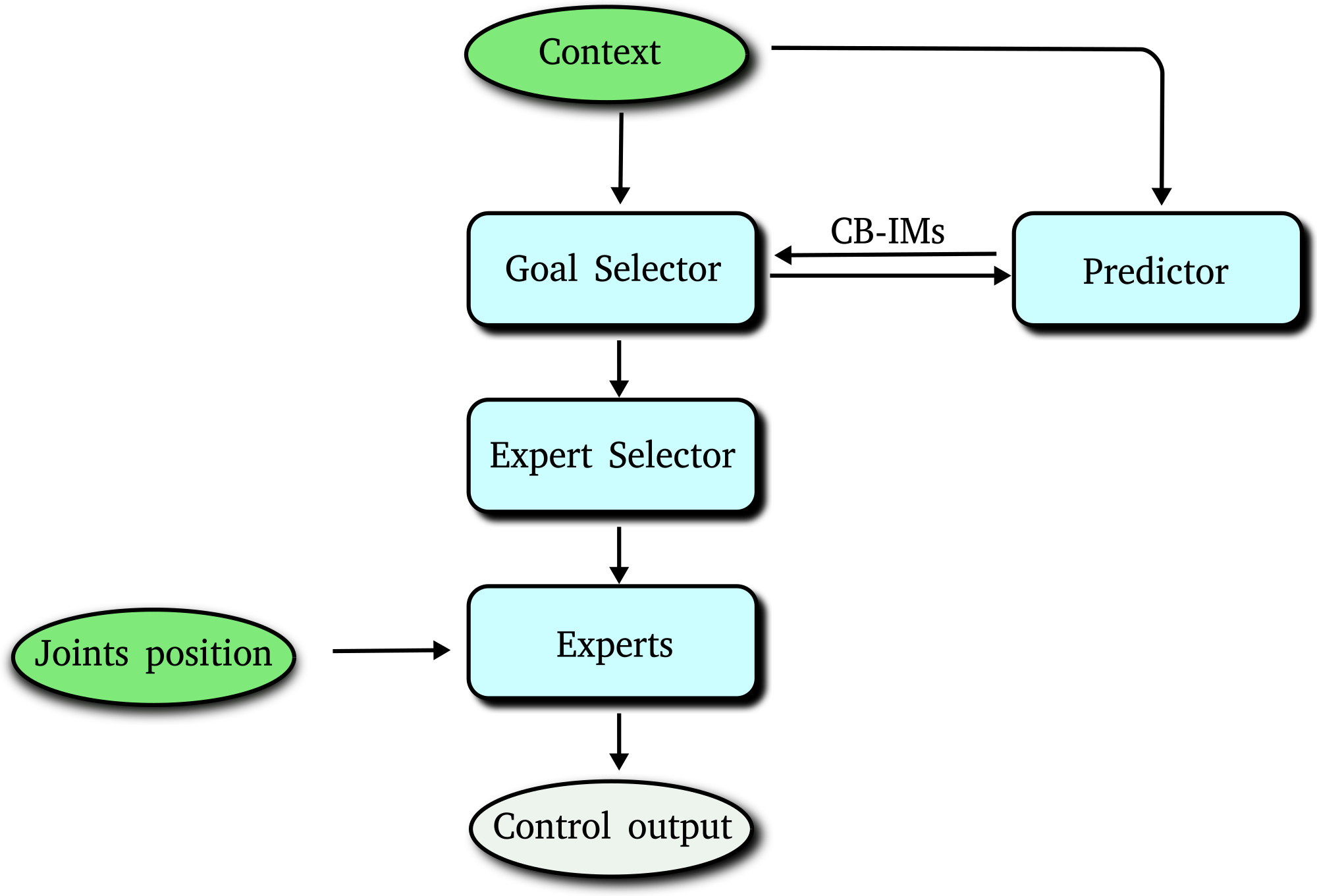}
    \caption{A schema of the architecture implemented in C-GRAIL and M-GRAIL. Differently from GRAIL, the new architectures use context as input to the goal selector. Note that for all the architectures the expert selector and experts are goal-specific.}
    \label{fig:Architecture}
\end{figure}

As described in sec. \ref{sec:ProblemDescription} the core of the autonomous learning process resides in the goal selection process happening in the GS. In the original version of GRAIL the GS receives no input and gives as output the selected goal as in a standard bandit setting, where each arm/goal is evaluated on the basis of an exponential moving average (EMA), with smoothing factor set to 0.01, of the previously-achieved intrinsic rewards (the CPI). Here we will present two different versions of GRAIL (see Fig. \ref{fig:Architecture} for a general schema of the new architectures) that, by modifying the GS component, are able to cope with the added complexity of the scenarios described in Sec. \ref{sec:Setup}. The first, called \textit{Contextual}-GRAIL (C-GRAIL) provides as input to the GS the state of the environment, which can be composed of standard state features or also features describing the status of different goals (e.g features describing whether each sphere is activated). The GS then selects the tasks as in a contextual bandit where different EMAs (with smoothing factor set to $0.1$) are associated with different contexts, and with the same \textit{softmax} selection rule as in GRAIL. A second version, called \textit{Markovian}-GRAIL (M-GRAIL), provides the same input to the GS as C-GRAIL, but treats goal selection as a reinforcement learning MDP and solves it by modeling the temporal interdependency between goals as the temporal dependency between consecutive states in an MDP; it then uses Q-learning (with a learning rate of $0.1$ and a discount factor set to $0.3$) to assign a value to each goal. In particular, here values represent the long-term benefits of practicing a goal considering the intrinsic rewards that other goals that depend on it may provide in the future. Goal selection follows the same \textit{softmax} selection rule as the previous system. 

Moreover, C-GRAIL and M-GRAIL use the information on the status of the spheres (the contextual input): \textit{(a)} to generate condition-specific CB-IMs for each goal and \textit{(b)} to avoid a ``disruptive'' training of the low-level policies. Mechanism \textit{(a)} is implemented by providing the contextual input also to the predictor that generates the CPI signal. Mechanism \textit{(b)} is implemented by blocking the learning of the selected expert when the predictor of the selected goal has a $0$ output, unless the goal is achieved. This avoids situations where a goal is selected even if its preconditions are not satisfied: a trained expert would bring the robot on the sphere but there would be no effect (the sphere would not activate) and thus no reward signal, generating a ``disruptive'' modification of the policy. This ``error'' would not be due to an incorrect policy but to an improper selection of the GS. While the latter selection has to be ``punished'', the actor-critic does not have to modify its policy. Note that while in the original version of GRAIL mechanism \textit{(b)} could be implemented only using the knowledge of the designers,  C-GRAIL and M-GRAIL are able to autonomously regulate this process.

\section{Results}
\label{sec:Results}
This section presents the results of the three experiments described in Sec. \ref{sec:Setup}. All data are averages over ten replications of the tested systems. The experimental details of each scenario are presented at the beginning of the related sub-sections. 

\subsection{First experiment: no relations between tasks}
\label{sec:FirstExperiment}

\begin{figure}
    \centering
    \includegraphics[width=0.99 \columnwidth,keepaspectratio]{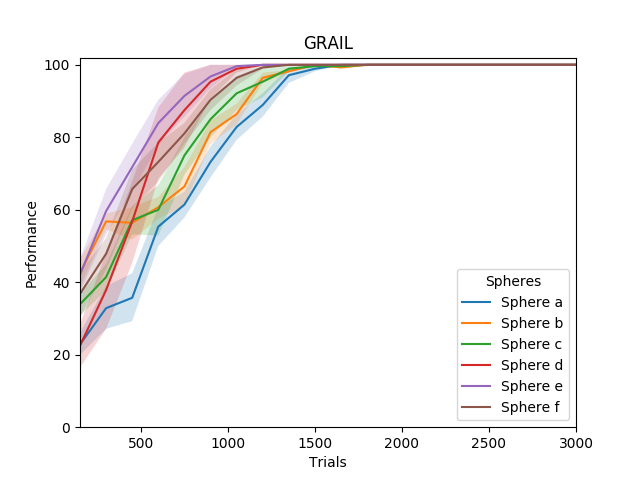}
    \caption{Performance of GRAIL in the first experiment. Average over 10 replications of the experiment. Shadows show the confidence intervals.}
    \label{fig:Exp1}
\end{figure}

This can be considered as the \textit{baseline} experiment, where we just show the performance of the original GRAIL in a scenario where the 6 tasks presented to the robot (learning to activate the 6 spheres) have no relations with the specific conditions of the environment, nor between each other. We run the experiment for 3000 trials, each one ending when the robot collides with one of the spheres or after a timeout of 800 steps. After each trial we reset the environment, i.e. the activated spheres are set to off.  At the beginning of each trial, the GS selects, as in a bandit problem, the task the system tries to achieve and then the related expert is trained. 

Fig.~\ref{fig:Exp1} shows how GRAIL is able to perfectly learn all the tasks in $\sim$1700 trials, properly shifting between them during the simulation thanks to the CB-IMs generated for the improvement of the different skills.

\subsection{Second experiment: environmental dependence}
\label{sec:SecondExperiment}

\begin{figure}
  \begin{center}
    \begin{minipage}{.49\textwidth}
       \centering
       \includegraphics[width=1\columnwidth]{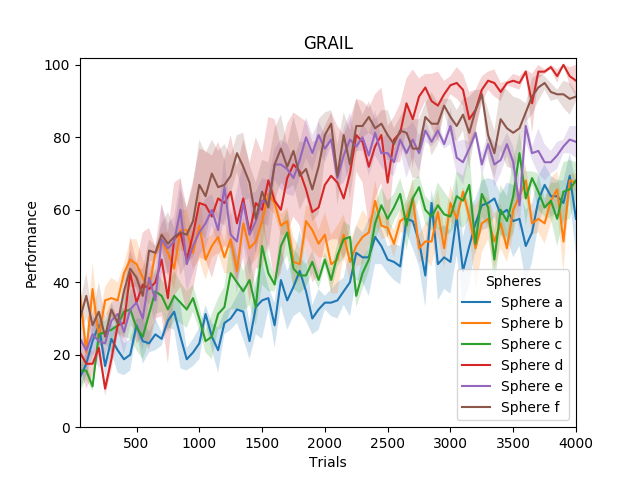}
    \end{minipage}
    \\
    \begin{minipage}{.49\textwidth}
       \centering
       \includegraphics[width=1\columnwidth]{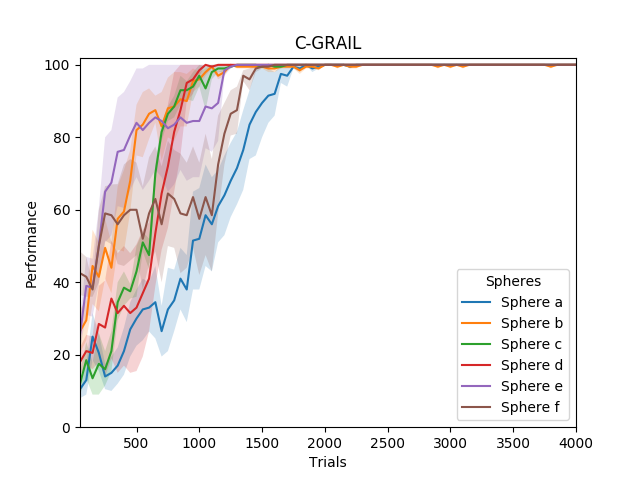}
    \end{minipage}
    \end{center}
    \caption{Performance of GRAIL and C-GRAIL in the second experiment.}
    \label{fig:Exp2}
\end{figure}

\begin{figure}
  \begin{center}
    \begin{minipage}{.49\columnwidth}
       \centering
       \includegraphics[width=1\columnwidth]{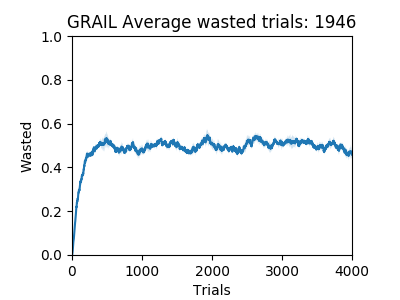}
    \end{minipage}
    \begin{minipage}{.49\columnwidth}
       \centering
       \includegraphics[width=1\columnwidth]{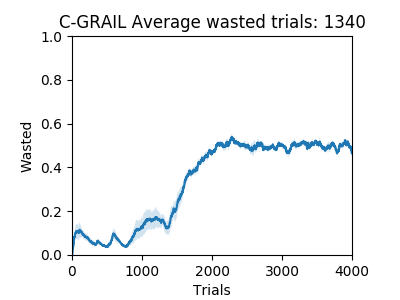}
    \end{minipage}
    \end{center}
    \caption{Trials wasted by GRAIL and C-GRAIL in selecting tasks that cannot be performed.}
    \label{fig:Exp2_Wasted}
\end{figure}

In this second experiment we compare GRAIL and C-GRAIL in a modified setup where the value of a contextual feature ($cf$) is used as precondition to determine whether the agent can activate certain spheres: spheres $a$, $c$ and $e$ can be activated only when the $cf$ is set to 1.0, while spheres $b$, $d$ and $f$ only when $cf$ is set to 0.0. At the beginning of each trial, $cf$ is set to 1.0 with 50\% probability. While GRAIL selects tasks without considering the environmental condition, C-GRAIL receives $cf$ status as input and performs task selection as in a contextual bandit.

Fig.~\ref{fig:Exp2} shows the performances of GRAIL and C-GRAIL on the 6 tasks during the experiment that lasted 4000 trials. C-GRAIL is able to properly learn all the tasks in $\sim$2000 trials, while at the end of the simulation GRAIL has reached a high competence (over 80\%) only in 2 tasks. This because GRAIL performs goal evaluation and selection without considering the status of the $cf$ which instead is decisive for spheres activation. As shown by Fig. \ref{fig:Exp2_Wasted}, while C-GRAIL only wastes trials in ``random'' selection when all the tasks are properly learnt, this meaning that it is able to generate an IM for the different tasks only in the conditions where they can be actually achieved, GRAIL wastes time from the beginning of the experiment in selecting tasks also when they cannot be trained, thus impairing the learning process.

\subsection{Third experiment: multiple interrelated tasks}
\label{sec:ThirdExperiment}

\begin{figure}
    \centering
    \includegraphics[width=0.3 \columnwidth,keepaspectratio]{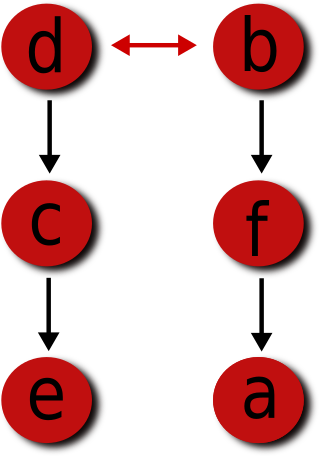}
    \caption{Structure of the third experimental scenario. Black arrows indicate positive dependencies, and red arrows negative dependencies.}
    \label{fig:Exp3_Structure}
\end{figure}

The third experiment increases the complexity of the scenario with the introduction of interrelations between tasks. This allows us to test the validity of our main hypothesis, implemented in the M-GRAIL system. In particular, in this experiment we have two ``chains'' of interdependent tasks (see Fig.~\ref{fig:Exp3_Structure}): spheres $d \rightarrow c \rightarrow e$ and $b \rightarrow f \rightarrow a$ (where arrows indicate that a sphere is the precondition for the following one). Moreover spheres $d$ and $b$, at the beginning of the two sequences, are mutually exclusive, so that if the robot starts one chain it cannot turn on the spheres of the other. Since here we focus on task-interdependencies and sequences, the scheduling is as follows: each simulation is run for 2000 ``epochs'', where each epoch lasts 3 trials (6000 trials in total). At the end of each epoch we reset the environment, while during the epoch the spheres remain in the status determined by the activity of the robot. 

\begin{figure}
  \begin{center}
    \begin{minipage}{.49\textwidth}
       \centering
       \includegraphics[width=1\columnwidth]{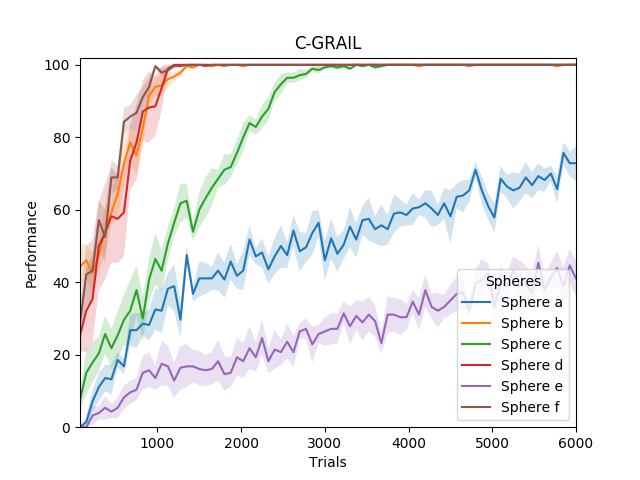}
    \end{minipage}
    \\
    \begin{minipage}{.49\textwidth}
       \centering
       \includegraphics[width=1\columnwidth]{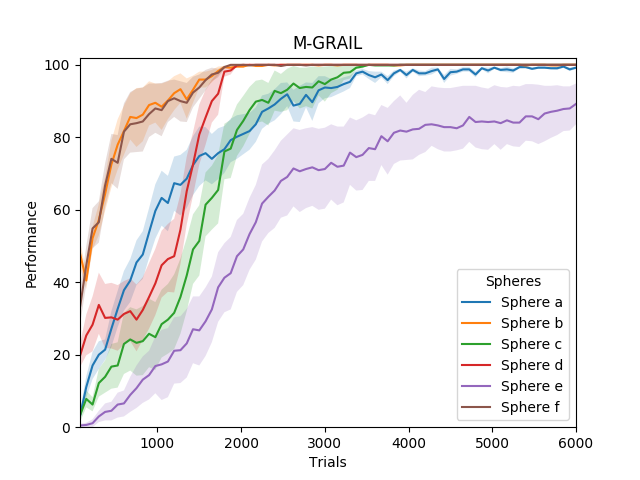}
    \end{minipage}
    \end{center}
    \caption{Performance of C-GRAIL and M-GRAIL in the third experiment.}
    \label{fig:Exp3}
\end{figure}

\begin{figure}
  \begin{center}
    \begin{minipage}{.49\columnwidth}
       \centering
       \includegraphics[width=1\columnwidth]{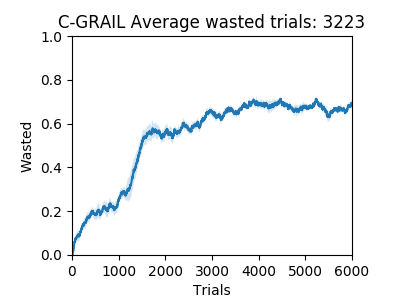}
    \end{minipage}
    \begin{minipage}{.49\columnwidth}
       \centering
       \includegraphics[width=1\columnwidth]{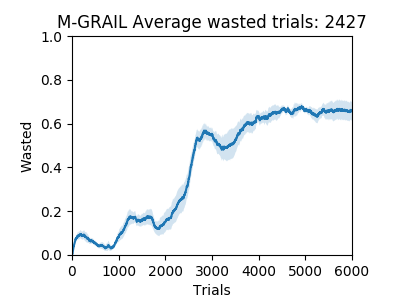}
    \end{minipage}
    \end{center}
    \caption{Trials wasted by C-GRAIL and M-GRAIL in selecting tasks that cannot be performed.}
    \label{fig:Exp3_Wasted}
\end{figure}

From the second experiment (Sec.~\ref{sec:SecondExperiment}) we know that GRAIL system is not able to properly perform autonomous learning when tasks are dependent on some preconditions, so here we tested only C-GRAIL and M-GRAIL. Both systems get the on/off status of the six spheres as input to the goal selector, but they implement value assignment in the two different ways explained in Sec.~\ref{sec:M-GRAIL}.  

The performance of the two systems is presented in Fig.~\ref{fig:Exp3}. In this scenario C-GRAIL can properly learn 4 tasks but it is not able  to achieve high competence on spheres $a$ and $e$, the last ones of the two chains. Instead, M-GRAIL reach a high performance in all the skills: 5 over 6 are completely learnt during the simulations and sphere $e$ reaches a performance close to 90\% (on average over the 10 replications). Although both systems are able to assign values to goals only in the states where their preconditions are satisfied, C-GRAIL suffers from the fact that some tasks are ``farther'' from the initial condition (all spheres off). Whenever a task is completely learnt (the robot has an optimal policy for performing a goal), the intrinsic motivation for selecting it gradually disappears. This may lead to a situation where the robot starts selecting tasks almost at random due to the absence of intrinsic rewards, thus wasting trials in selecting goals that cannot be achieved at that moment (Fig. \ref{fig:Exp3_Wasted}). As a result, even though the robot may have an intrinsic motivation reward for practicing a goal (e.g. activating sphere $a$ and $e$), it does not have intrinsic motivation for first practicing the goals that are preconditions to $a$ and to $e$; it is not, thus, capable of systematically putting the environment in the proper conditions to train the last spheres of the chains. On the contrary, M-GRAIL can rapidly learn all tasks: even when (similarly to C-GRAIL) it is no longer intrinsically motivated in achieving ``simple'' goals \textit{per se} (i.e., goals with few preconditions), it ensures that the robot continues to select those goals thanks to the Q-learning algorithm, which propagates the intrinsic motivations for solving task $a$ and $e$ back to the tasks that are their preconditions. Thanks to this strategy in goal selection, M-GRAIL starts wasting trials only when all the goals have reached a high performance (Fig. \ref{fig:Exp3_Wasted}). 

\section{Conclusions}
\label{sec:Conclusions}

In this work we tackled the crucial problem of autonomous learning of multiple interrelated tasks in robotic scenarios. As described in Sec.~\ref{sec:ProblemDescription}, our hypothesis is that open-ended learning of skills should be treated as a problem of active task selection to be solved with different strategies given the structure of the experimental scenario. In particular, when the tasks are independent from environmental conditions and from each other, a classical $N$-armed bandit strategy is sufficient to properly train the low-level skills related to each task (Sec.~\ref{sec:FirstExperiment}). When the tasks are conditioned to specific conditions of the environment independent from robot activity, treating the problem as a $contextual$ bandit allows a proper learning of the skills (Sec.~\ref{sec:SecondExperiment}). But if we want to autonomously learn sequences of interdependent tasks, we have to consider task selection as an MDP where the choices of the system determine changes in the state of the environment (sec \ref{sec:ThirdExperiment}). In this way the system can use algorithms such as Q-learning to transfer the intrinsic-motivation value for learning a task to the related ones, thus allowing hierarchical skill learning. 

This is particularly important for open-ended learning scenarios, where task-agnostic motivation signals are used to drive goal selection and skill learning. These heuristics are perfect to smartly shift from one task to another, since the intrinsic value of a goal remains only when the system has something to learn and disappears otherwise. However, this could be a problem if a task constitutes the precondition for another one, so that the robot needs to perform the first even if the motivation for it has faded out. A system such as the presented M-GRAIL is able to keep the advantages of intrinsically motivated learning and at the same time to cope with the crucial problem of multiple interrelated tasks learning. Kulkarni and colleagues \cite{Kulkarni2016} presented an interesting model combining hierarchical deep reinforcement learning and IMs in a two levels architecture similar to M-GRAIL, where a high-level meta-learner is selecting goals and a lower lever controller is learning the policies to achieve those goals. However, in their work the higher level is maximising extrinsic rewards (achieving goals) while intrinsic motivations are used for the lower levels: differently from what we are tackling in our study, this solution leads a system to focus on those goals that are providing more rewards instead of making the system learning many different goals.
 
In all the different versions of GRAIL the experts that acquire the skills are implemented as actor-critic neural networks. However, any other more efficient method can be used, e.g. parametrised skills such as Dynamic Movement Primitives (or variations of them) trained through policy search algorithms \cite{Schaal2005}. The generalisation of the acquired skills over new/different targets was not the focus of our work. Using different kind of controllers might facilitate this process, as well as leveraging on visual input to guide motor behaviour and to perform target recognition \cite{Sperati2017}.

A stronger limitation of M-GRAIL resides in the fact that although it can select (and learn) hierarchical tasks, the system is not able to retain these ``chains'' after the learning process, i.e. to be able to select and perform them as whole skills. Modifying the structure of the experts could provide a solution to this limitation, as well as considering the implementation of a planner on top of M-GRAIL to compose sequences of the autonomously acquired skills using higher-level encoding \cite{Konidaris2018}.

\section*{Acknowledgment}
This project has received funding from the European Union's Horizon 2020 Research and Innovation Programme under Grant Agreement no 307010 (GOAL-Robots - Goal-based Open-ended Autonomous Learning Robots).

\bibliographystyle{IEEEtran}
\bibliography{Ref_Icdl2019}

\begin{thebibliography}{10}
\providecommand{\url}[1]{#1}
\csname url@samestyle\endcsname
\providecommand{\newblock}{\relax}
\providecommand{\bibinfo}[2]{#2}
\providecommand{\BIBentrySTDinterwordspacing}{\spaceskip=0pt\relax}
\providecommand{\BIBentryALTinterwordstretchfactor}{4}
\providecommand{\BIBentryALTinterwordspacing}{\spaceskip=\fontdimen2\font plus
\BIBentryALTinterwordstretchfactor\fontdimen3\font minus
  \fontdimen4\font\relax}
\providecommand{\BIBforeignlanguage}[2]{{%
\expandafter\ifx\csname l@#1\endcsname\relax
\typeout{** WARNING: IEEEtran.bst: No hyphenation pattern has been}%
\typeout{** loaded for the language `#1'. Using the pattern for}%
\typeout{** the default language instead.}%
\else
\language=\csname l@#1\endcsname
\fi
#2}}
\providecommand{\BIBdecl}{\relax}
\BIBdecl

\bibitem{Oudeyer2007intrinsic}
P.-Y. Oudeyer, F.~Kaplan, and V.~Hafner, ``Intrinsic motivation systems for
  autonomous mental development,'' \emph{IEEE transactions on evolutionary
  computation}, vol.~11, no.~6, 2007.

\bibitem{Schmidhuber2010}
J.~Schmidhuber, ``Formal theory of creativity, fun, and intrinsic motivation
  (1990--2010),'' \emph{IEEE Transactions on Autonomous Mental Development},
  vol.~2, no.~3, pp. 230--247, 2010.

\bibitem{Baldassarre2013Book}
G.~Baldassarre and M.~Mirolli, \emph{Intrinsically Motivated Learning in
  Natural and Artificial Systems}.\hskip 1em plus 0.5em minus 0.4em\relax
  Springer Science \& Business Media, 2013.

\bibitem{Santucci2014}
V.~G. Santucci, G.~Baldassarre, and M.~Mirolli, ``Cumulative learning through
  intrinsic reinforcements,'' in \emph{Evolution, Complexity and Artificial
  Life}.\hskip 1em plus 0.5em minus 0.4em\relax Springer, 2014, pp. 107--122.

\bibitem{Hafez2017}
M.~B. Hafez, C.~Weber, and S.~Wermter, ``Curiosity-driven exploration enhances
  motor skills of continuous actor-critic learner,'' in \emph{Proceedings of
  the 7th Joint IEEE International Conference on Development and Learning and
  Epigenetic Robotics (ICDL-EpiRob)}, 2017.

\bibitem{Dhakan2018}
P.~Dhakan, K.~Merrick, I.~Ra{\~n}{\'o}, and N.~Siddique, ``Intrinsic rewards
  for maintenance, approach, avoidance, and achievement goal types,''
  \emph{Frontiers in neurorobotics}, vol.~12, 2018.

\bibitem{Tanneberg2019}
D.~Tanneberg, J.~Peters, and E.~Rueckert, ``Intrinsic motivation and mental
  replay enable efficient online adaptation in stochastic recurrent networks,''
  \emph{Neural Networks}, vol. 109, pp. 67--80, 2019.

\bibitem{Baranes2013}
A.~Baranes and P.-Y. Oudeyer, ``Active learning of inverse models with
  intrinsically motivated goal exploration in robots,'' \emph{Robotics and
  Autonomous Systems}, vol.~61, no.~1, pp. 49--73, 2013.

\bibitem{Santucci2016}
V.~G. Santucci, G.~Baldassarre, and M.~Mirolli, ``Grail: A goal-discovering
  robotic architecture for intrinsically-motivated learning,'' \emph{IEEE
  Transactions on Cognitive and Developmental Systems}, vol.~8, no.~3, pp.
  214--231, 2016.

\bibitem{Forestier2017}
S.~Forestier, Y.~Mollard, and P.-Y. Oudeyer, ``Intrinsically motivated goal
  exploration processes with automatic curriculum learning,'' \emph{arXiv
  preprint arXiv:1708.02190}, 2017.

\bibitem{Lopes2012}
M.~Lopes and P.-Y. Oudeyer, ``The strategic student approach for life-long
  exploration and learning,'' in \emph{Development and Learning and Epigenetic
  Robotics (ICDL), 2012 IEEE International Conference on}.\hskip 1em plus 0.5em
  minus 0.4em\relax IEEE, 2012, pp. 1--8.

\bibitem{Santucci2013best}
V.~G. Santucci, G.~Baldassarre, and M.~Mirolli, ``Which is the best intrinsic
  motivation signal for learning multiple skills?'' \emph{Frontiers in
  neurorobotics}, vol.~7, p.~22, 2013.

\bibitem{Seepanomwan2017}
K.~Seepanomwan, V.~G. Santucci, and G.~Baldassarre, ``Intrinsically motivated
  discovered outcomes boost user’s goals achievement in a humanoid robot,''
  in \emph{2017 Joint IEEE International Conference on Development and Learning
  and Epigenetic Robotics (ICDL-EpiRob)}, 2017, pp. 178--183.

\bibitem{Barto2003}
A.~G. Barto and S.~Mahadevan, ``Recent advances in hierarchical reinforcement
  learning,'' \emph{Discrete event dynamic systems}, vol.~13, no. 1-2, pp.
  41--77, 2003.

\bibitem{Vigorito2010}
C.~M. Vigorito and A.~G. Barto, ``Intrinsically motivated hierarchical skill
  learning in structured environments,'' \emph{IEEE Transactions on Autonomous
  Mental Development}, vol.~2, no.~2, pp. 132--143, 2010.

\bibitem{Bakker2004}
B.~Bakker and J.~Schmidhuber, ``Hierarchical reinforcement learning based on
  subgoal discovery and subpolicy specialization,'' in \emph{Proc. of the 8-th
  Conf. on Intelligent Autonomous Systems}, 2004, pp. 438--445.

\bibitem{Niel2018}
R.~Niel and M.~A. Wiering, ``Hierarchical reinforcement learning for playing a
  dynamic dungeon crawler game,'' in \emph{2018 IEEE Symposium Series on
  Computational Intelligence (SSCI)}.\hskip 1em plus 0.5em minus 0.4em\relax
  IEEE, 2018, pp. 1159--1166.

\bibitem{Reinhart2017}
R.~F. Reinhart, ``Autonomous exploration of motor skills by skill babbling,''
  \emph{Autonomous Robots}, vol.~41, no.~7, pp. 1521--1537, 2017.

\bibitem{DaSilva2014}
B.~C. Da~Silva, G.~Baldassarre, G.~Konidaris, and A.~Barto, ``Learning
  parameterized motor skills on a humanoid robot,'' in \emph{Robotics and
  Automation (ICRA), 2014 IEEE International Conference on}.\hskip 1em plus
  0.5em minus 0.4em\relax IEEE, 2014, pp. 5239--5244.

\bibitem{Grollman2010}
D.~H. Grollman and O.~C. Jenkins, ``Incremental learning of subtasks from
  unsegmented demonstration,'' in \emph{Intelligent Robots and Systems (IROS),
  2010 IEEE/RSJ International Conference on}.\hskip 1em plus 0.5em minus
  0.4em\relax IEEE, 2010, pp. 261--266.

\bibitem{Mohseni2018}
A.~Mohseni-Kabir, C.~Li, V.~Wu, D.~Miller, B.~Hylak, S.~Chernova, D.~Berenson,
  C.~Sidner, and C.~Rich, ``Simultaneous learning of hierarchy and primitives
  for complex robot tasks,'' \emph{Autonomous Robots}, pp. 1--16, 2018.

\bibitem{Nair2018}
A.~Nair, B.~McGrew, M.~Andrychowicz, W.~Zaremba, and P.~Abbeel, ``Overcoming
  exploration in reinforcement learning with demonstrations,'' in \emph{2018
  IEEE International Conference on Robotics and Automation (ICRA)}.\hskip 1em
  plus 0.5em minus 0.4em\relax IEEE, 2018, pp. 6292--6299.

\bibitem{Duminy2018}
N.~Duminy, S.~M. Nguyen, and D.~Duhaut, ``Learning a set of interrelated tasks
  by using a succession of motor policies for a socially guided intrinsically
  motivated learner,'' \emph{Frontiers in neurorobotics}, vol.~12, 2018.

\bibitem{Rolf2014}
M.~Rolf and M.~Asada, ``Autonomous development of goals: From generic rewards
  to goal and self detection,'' in \emph{Development and Learning and
  Epigenetic Robotics (ICDL-Epirob), 2014 Joint IEEE International Conferences
  on}.\hskip 1em plus 0.5em minus 0.4em\relax IEEE, 2014, pp. 187--194.

\bibitem{Meeden2017}
L.~Meeden and D.~Blank, ``Developing grounded goals through instant replay
  learning,'' in \emph{The Seventh Joint IEEE International Conference on
  Development and Learning and on Epigenetic Robotics}, 2017.

\bibitem{Cartoni2018}
E.~Cartoni and G.~Baldassarre, ``Autonomous discovery of the goal space to
  learn a parameterized skill,'' \emph{arXiv preprint arXiv:1805.07547}, 2018.

\bibitem{Sutton1998}
R.~S. Sutton and A.~G. Barto, \emph{Reinforcement learning: An
  introduction}.\hskip 1em plus 0.5em minus 0.4em\relax MIT press, 1998.

\bibitem{Florensa2018}
C.~Florensa, D.~Held, X.~Geng, and P.~Abbeel, ``Automatic goal generation for
  reinforcement learning agents,'' in \emph{Proceedings of the 35th
  International Conference on Machine Learning}, ser. Proceedings of Machine
  Learning Research, J.~Dy and A.~Krause, Eds., vol.~80.\hskip 1em plus 0.5em
  minus 0.4em\relax Stockholmsmässan, Stockholm Sweden: PMLR, 10--15 Jul 2018,
  pp. 1514--1523.

\bibitem{Kulkarni2016}
T.~D. Kulkarni, K.~Narasimhan, A.~Saeedi, and J.~Tenenbaum, ``Hierarchical deep
  reinforcement learning: Integrating temporal abstraction and intrinsic
  motivation,'' in \emph{Advances in neural information processing systems},
  2016, pp. 3675--3683.

\bibitem{Santucci2013iccm}
V.~G. Santucci, G.~Baldassarre, and M.~Mirolli, ``Intrinsic motivation signals
  for driving the acquisition of multiple tasks: a simulated robotic study,''
  in \emph{Proceedings of the 12th International Conference on Cognitive
  Modelling (ICCM)}, 2013.

\bibitem{Merrick2012}
K.~E. Merrick, ``Intrinsic motivation and introspection in reinforcement
  learning,'' \emph{IEEE Transactions on Autonomous Mental Development},
  vol.~4, no.~4, pp. 315--329, 2012.

\bibitem{Levine2017}
N.~Levine, K.~Crammer, and S.~Mannor, ``Rotting bandits,'' in \emph{Advances in
  Neural Information Processing Systems}, 2017, pp. 3074--3083.

\bibitem{Metta2008}
G.~Metta, G.~Sandini, D.~Vernon, L.~Natale, and F.~Nori, ``The icub humanoid
  robot: an open platform for research in embodied cognition,'' in
  \emph{Proceedings of the 8th workshop on performance metrics for intelligent
  systems}.\hskip 1em plus 0.5em minus 0.4em\relax ACM, 2008, pp. 50--56.

\bibitem{Santucci2014icdl}
V.~G. Santucci, G.~Baldassarre, and M.~Mirolli, ``Autonomous selection of the
  “what” and the “how” of learning: an intrinsically motivated system
  tested with a two armed robot,'' in \emph{Development and Learning and
  Epigenetic Robotics (ICDL-Epirob), 2014 Joint IEEE International Conferences
  on}.\hskip 1em plus 0.5em minus 0.4em\relax IEEE, 2014, pp. 434--439.

\bibitem{Doya2000}
K.~Doya, ``Reinforcement learning in continuous time and space,'' \emph{Neural
  computation}, vol.~12, no.~1, pp. 219--245, 2000.

\bibitem{Schaal2005}
S.~Schaal, J.~Peters, J.~Nakanishi, and A.~Ijspeert, ``Learning movement
  primitives,'' in \emph{Robotics research. the eleventh international
  symposium}.\hskip 1em plus 0.5em minus 0.4em\relax Springer, 2005, pp.
  561--572.

\bibitem{Sperati2017}
V.~Sperati and G.~Baldassarre, ``Bio-inspired model learning visual goals and
  attention skills through contingencies and intrinsic motivations,''
  \emph{IEEE Transactions on Cognitive and Developmental Systems}, vol.~10,
  no.~2, pp. 326--344, 2017.

\bibitem{Konidaris2018}
G.~Konidaris, L.~P. Kaelbling, and T.~Lozano-Perez, ``From skills to symbols:
  Learning symbolic representations for abstract high-level planning,''
  \emph{Journal of Artificial Intelligence Research}, vol.~61, pp. 215--289,
  2018.

\end{thebibliography}

\end{document}